\documentclass{article}
\usepackage{arxiv}
\usepackage{graphicx}
\usepackage{subfig}
\usepackage{tcolorbox}
\usepackage{xcolor}
\usepackage{booktabs}
\usepackage{graphicx}
\usepackage{color, colortbl}
\usepackage{times}
\usepackage{soul}
\usepackage{url}
\usepackage{multirow}
\usepackage[hidelinks]{hyperref}
\usepackage[utf8]{inputenc}
\usepackage{caption}
\usepackage{graphicx}
\usepackage{amsmath}
\usepackage{amssymb}
\usepackage[noend]{algpseudocode}
\usepackage{booktabs}
\usepackage{algorithm}
\usepackage{xcolor}
\usepackage{tikz}
\usetikzlibrary{positioning}
\usepackage{pgfplots}
\definecolor{Gray}{gray}{0.9}

\title{Deep Knowledge-infusion for Explainable
Depression Detection}

\author{Sumit Dalal$^1$, Sarika Jain$^1$, and Mayank Dave$^1$\\
$^1$National Institute of Technology Kurukshetra, India \\
}

\renewcommand{\shorttitle}{\textit{Security and Privacy-Preservation of IoT Data in Cloud-Fog Computing Environment}}

\hypersetup{
pdfsubject={q-bio.NC, q-bio.QM},
pdfauthor={David S.~Hippocampus, Elias D.~Striatum},
pdfkeywords={First keyword, Second keyword, More},
}

\begin{document}

\maketitle

\begin{abstract}
Discovering individuals’ depression on social media has become increasingly important. Researchers employed ML/DL or lexicon-based methods for automated depression detection. Lexicon-based methods, explainable and easy to implement, match words from user posts in a depression dictionary without considering contexts, and little attention has been paid to how the word can be associated with the depression-related context. While the DL models can leverage contextual information, their black-box nature limits their adoption in the domain. Though surrogate models like LIME and SHAP can produce explanations for DL models, the explanations are suitable for the developer and of limited use to the end user. We propose a Knolwedge-infused Neural Network (KiNN) incorporating domain-specific knowledge from DepressionFeature ontology (DFO) in a neural network to endow the model with user-level explainability regarding concepts and processes the clinician understands. Further, commonsense knowledge from the Commonsense Transformer (COMET) trained on ATOMIC is also infused to consider the generic emotional aspects of user posts in depression detection. The model is evaluated on three expertly curated datasets related to depression. We observed the model to have a statistically significant (p$<$0.1) boost in performance over the best domain-specific model, MentalBERT, across CLEF e-Risk (25\% MCC$\uparrow$, 12\%F1$\uparrow$). A similar trend is observed across the PRIMATE dataset where the proposed model performed better than MentalBERT (2.5\% MCC$\uparrow$, 19\%F1$\uparrow$). The observations confirm the generated explanations to be informative for MHPs compared to post-hoc model explanations. Results demonstrated that the user-level explainability of KiNN also surpasses the performance of baseline models and can provide explanations where other baselines fall short. Infusing the domain and commonsense knowledge in KiNN enhances the ability of models like GPT-3.5 to generate application-relevant explanations. 

\end{abstract}


\section{Introduction} \label{sec:introduction}

Large population suffering from various mental disorders join social communities, a popular means of online communication for sharing and helping others. People tend to write daily posts covering feelings, physical moments, food habits, exercise, and music choices. Information from these posts is considered for the user's mental health assessment. Researchers observed signs of depression in user's social data well before their first diagnosis. Moreover, the social data can be collected non-intrusively. These advantages help health professionals detect the user's mental health without much interference in the user's life. 



ML/DL or lexicon-based approaches are manipulated to analyze the big data generated on social media platforms for depression detection. In lexicon-based approaches, users' social posts are searched for terms from specific dictionaries (like antidepressant or depression-related). If the frequency of the terms from these dictionaries in the user's posts crosses a threshold, the user is assigned a depressed label. While applying the lexicon-based method has been known to be explainable and easy to implement \cite{kotelnikova2021lexicon, razova2021does}. Their possible disadvantage, however, is that they only consider if terms in user posts match the depression lexicon and don't consider how a word can be associated with the depression-related context. For example, Figure 1 illustrates two sentences: “I cut my wrist” and “I have my hair cut.” Assuming that the word ‘cut’ belongs to the depression dictionary, only the former sentence should be evaluated as having depression. However, the latter sentence could also have depression if the methods of prior work \cite{ghosh2021depression, pan2023umuteam} are applied. In other words, if the context is incorrectly captured, a model using a depression lexicon created by experts may not be able to accurately assess the risk of depression \cite{limsopatham2016normalising}. 

While the ML/DL models can leverage the contextual information, their black-box nature limits adoption in the domain \cite{nadeem2023pewresearch}. They fail to provide explanations grounded in the knowledge that aligns with MHPs \cite{cushing2023health} and depend on surrogate models like LIME and SHAP to explain the decisions \cite{ribeiro2016should,lundberg2017unified}. However, these system-level explanations are helpful to developers and of limited use to the end user.

\begin{figure*}
    \centering
    \includegraphics[width=0.99\textwidth]{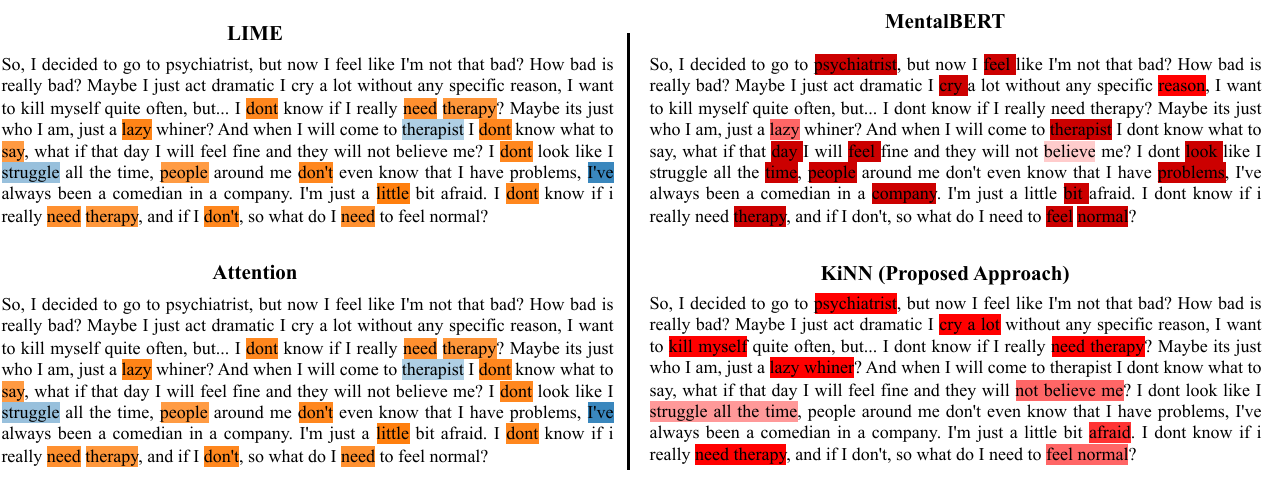}
    \caption{Comparision of explanations produced by LIME, MentalBERT, Self-Attention, and the proposed {\fontfamily{cmss}\selectfont KiNN} model. The objective is to enhance the ability to provide user-level explanations for depression detection. {\fontfamily{cmss}\selectfont KiNN} highlights variable-length phrases centered around depression, unlike others, which focus on unigrams lacking context for user understand-ability.}
    \label{fig:attentionvisualization}
\end{figure*}

Expert-level explainability can be defined as the connections between an AI model's collective experiences from training and the real-world entities and definitions that make sense to domain experts \cite{ehsan2021expanding}. Consider the following expression, ``\textit{For the past several weeks, I have no to little interest to write my life any better than it is at the moment.}'' is an example of ``little interest or pleasure in doing things'' which is the first question in Patient Health Questionnaire (PHQ-9), which is a clinical instrument for measuring the severity of depression. Phrases like \textit{for the past several weeks, no to little interest} are MHP-explainable concepts. If a user's post containing the above-mentioned sentence is classified as \textbf{depressed} by an AI model, then such phrases should be the focus. In this direction, \textit{Explainable AI} has gained attention in natural language processing (NLP) for mental health, as it provides ``explanations'' for the BlackBox model's decisions \cite{zirikly2022explaining}. The explanations in LIME and SHAP are obtained through training an interpretable classifier that matches the BlackBox's outputs. Examples of such BlackBox AI models are BERT, RoBERTa, Longformer, and other self-attention-based language models \cite{devlin2018bert}, \cite{liu2019roberta}, \cite{beltagy2020longformer}. The attention visualizations are limited to MHPs and require post-hoc explainability techniques, which have other issues \cite{bordt2022post}. Supporting BlackBox models with clinical knowledge, like PHQ-9, Diagnostic and Statistical Manual for Mental Health Disorders (DSM-5), which MHPs often use, would result in models capable of delivering \textit{\textbf{user-level explanations}}. Alternatively, recent studies have demonstrated the importance of clinical knowledge and expertise in creating labeled datasets that improve the quality of explanations from BlackBox models \cite{zirikly2022explaining}, \cite{gaur2019knowledge}. However, constructing such datasets with infused knowledge has issues with quality (e.g., agreement scores between MHPs), transferability, cost, and effort. Figure \ref{fig:attentionvisualization} displays significant features by various approaches considered for generating explanations. Existing methods (LIME, MentalBERT, and Attention) highlight unigrams most of which have low domain-specific meaning. However, the proposed approach can identify the domain-specific and bigger grams (bi or tri) compared to the previous ones. Bigger grams have more significance than smaller ones as they contain more contextual information. For domain-specific results, background knowledge is introduced in terms of the domain ontology.



Knowledge infusion is divided into three levels: a.) Shallow Infusion b.) Semi-Deep Infusion c.) Deep Infusion. In shallow infusion, the knowledge and the infusion method are shallow as syntactic and lexical knowledge is considered through word embedding models. While in semi-deep infusion, external knowledge is involved through attention mechanisms or learnable knowledge constraints acting as a sentinel to guide model learning. Further, in deep infusion, a stratified representation of knowledge representing different levels of abstraction is employed in different layers of a deep learning model to transfer knowledge that aligns with the corresponding layer in the layered learning process. 

Our main contributions are as follows:
\begin{enumerate}
    \item Machine learning and semantic technologies driven hybrid model is proposed for depression detection on the social media data. 
    \item Mental health-related contextual information is captured with domain-centric vector space representation of the n-grams and phrases in user posts.  
    \item Semantic information is retrieved from the advanced FeatureOnto ontology that adds depression-related conceptual phrases to the existing ontology version.  
    \item Classification results of the proposed approach are compared with state-of-the-art machine learning and deep learning methods.
\end{enumerate}

\section{Literature}

\subsection{Knowledge-infusion in Depression Detection}
Various aspects of a social media user, along with the user's social postings, have been utilized in detecting depression, for example, the social media user's profiles and his/her behavior on the social platform \cite{bucur2021early, dinu2021automatic}. The downside is that these models are trained on several irrelevant contents, which might not be crucial for detecting a depressed user. Besides, this content hurts the overall efficiency and effectiveness of the model. To improve the model performance, authors infused knowledge at various levels using dictionaries/lexicons, which were reported to be effective in capturing specific linguistic or domain-specific characteristics to detect depression \cite{zogan2020explainable, zogan2021depressionnet}. The authors of \cite{schoene2021hierarchical, zogan2021depressionnet, wang2021learning} mapped the words from users' textual posts into Lexicons like Linguistic Inquiry and Word Count (LIWC) \cite{tausczik2010psychological}, Affective norms for English words (ANEW) \cite{bradley1999affective}, and NRC \cite{mohammad2013crowdsourcing} for infusing linguistic knowledge in the classification process. While the authors of \cite{de2013c, song2018feature, trotzek2018word, trotzek2018utilizing, zogan2020explainable, zogan2021depressionnet} focussed on domain-specific knowledge like antidepressants. Few other works which considered domain-specific knowledge like exercise-related information are \cite{coppersmith2014measuring}, and disease or symptoms-related information are \cite{yazdavar2017semi, briand2018analysis, trotzek2018word, trotzek2018utilizing}.

\subsection{Explainable Depression Detection}
Sometimes, the explanation is confused with the interpretation work; however, both are closely related but not the same. Like in \cite{cai2023depression}, authors explored and interpreted multi-variate time series features for depressed versus non-depressed users in depression detection. For this purpose, they exploited the descriptive text of the depression symptoms to check the similarity index between each user post and the symptoms to calculate different time series features for a user. Here, the time-series features are interpreted, but the reason for the classification output is not produced. 

Authors of \cite{ive2018hierarchical} provided a glimpse of the role of attention weights in explaining depression classification results by hierarchical attention neural model. In \cite{song2018feature}, multiple features besides the similarity index of user posts with depression symptoms and antidepressants have been considered for depression detection from Reddit posts. The relation between the features and the representations learned by the post-level attention network is studied to interpret the depression classification results. Similarly, the authors of \cite{wongkoblap2021deep} emphasized post-level attention weights from hierarchical attention networks for producing depression classification explanations. The authors in \cite{kshirsagar2017detecting} attempted to generate explanations using representation learned by attention mechanism only on the words in a user post, with accompanying limitations. Also, in  \cite{sekulic2020adapting}, word-level attention representations from hierarchical attention networks are inspected to analyze unigrams and bigrams relevant for classification. However, the authors of \cite{zogan2022explainable} visualized the significant words and posts based on the attention weights from a hierarchical attention network of word and post levels to explain the classification results. They employed multiple features, including symptoms and antidepressant count, in depression detection from social media posts. Though a BiLSTM-CNN with Attention Network is provided for depression classification in \cite{ghosh2023attention}, they considered the word frequency of depressive and non-depressive samples from the training set to find feature-level explanations. These works focus on attention weights, but authors in \cite{burdisso2020tau} designed a new classifier. They updated the SS-3 classifier to accommodate grams of various lengths instead of uni-grams only from training samples for depression detection. The classifier works on a decision tree where n-gram nodes are given some weight depending on their participation in the classification process. All works mentioned here are considered user posts for explanation except the work in \cite{tong2022cost}. Numerical features like the number of followers, positive/negative emotion words, and posts have been deployed for explanation generation by TreeSHAP, a game approach for explaining decision tree outputs.

From the literature, we observed that only a few authors focussed on explanation generation regarding depression classification output. Some worked on redesigning the depression classifier to produce explainable output, while others concentrated on attention networks for explanation generation. However, none of these produced the user-level explanations our model tends to do. Moreover, the lexicons used for knowledge infusion lack domain-specific information incorporated at shallow or semi-deep levels. Further, prior work focused on matching a word in a post with a depression dictionary without considering contexts, and little attention has been paid to how the word can be associated with the depression-related context.

\section{\textbf{K}nowledge-\textbf{i}nfused \textbf{N}eural \textbf{N}etwork (KiNN)}

\begin{table}
\centering
\caption{Symbols Table}
\label{tab:KiNN_symbol_table}
\begin{tabular}{|p{0.75cm}| p{6.75cm}|} 
\hline
\textbf{Symbol} & \textbf{Abbreviation} \\
\hline
$u_i$ & $ith$ social media user's posts \\
\hline
$z_{u_i}$ & contextualized representation of $ith$ user's posts returned by {\fontfamily{cmss}\selectfont KiNN} \\
\hline
$u^{pt}_i$ & phrase tagged posts from $ith$ user \\
\hline
$\hat{u}^{pt}_i$ & encoded phrase tagged posts \\
\hline
$W$ & Writer or Speaker of the post/comment \\
\hline
$L$ & Listeners (others)\\
\hline
$E_{IW}$ & Intent of $W$ \\
\hline
$E_{EW}$ & Effect on $W$ \\
\hline
$E_{RW}$ & Reaction of $W$ \\
\hline
$E_{EL}$ & Effect on $L$ \\
\hline
$E_{RL}$ & Reaction on $L$ \\
\hline
$W$ & Noun token \\
\hline
$l$ & Noumber of noun tokens in user post $u_i$ \\
\hline
\end{tabular}
\end{table}

\subsection{Architecture}
\begin{figure*}
    \centering
    \includegraphics[width=0.89\textwidth]{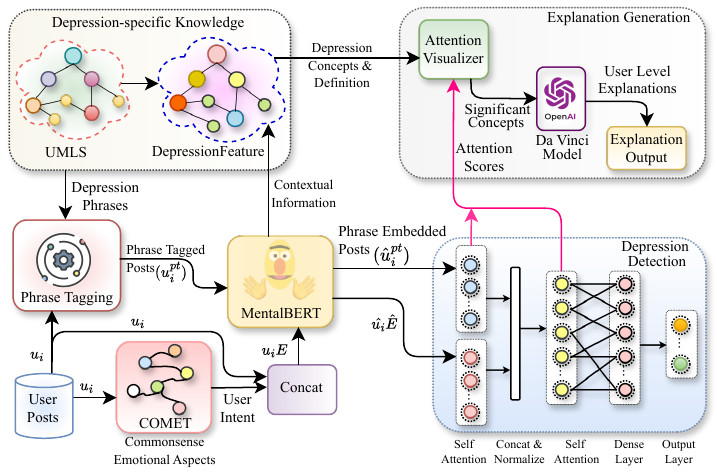}
    \caption{\footnotesize {\fontfamily{cmss}\selectfont \textbf{K}nowledge-\textbf{i}nfused \textbf{N}eural \textbf{N}etwork (KiNN)} model considering different aspects of user posts-domain and commonsense for depression detection. MentalBERT provides context-specific embeddings for phrase-tagged user posts. COMET provides nine aspects related to user posts, five of which are mental-health related and hence considered to check the intent of user posts. {\fontfamily{cmss}\selectfont KiNN} allows visualization of various attention layers for user-level explanations.}
    \label{fig:KiNN}
\end{figure*}

{\fontfamily{cmss}\selectfont KiNN} is a deep neural model with domain and commonsense knowledge infusion through multiple dense and attention layers. {\fontfamily{cmss}\selectfont KiNN} has a unique aspect of three-stage knowledge infusion: shallow infusion via domain-specific embeddings, semi-deep infusion via knowledge incorporation during neural network weight learning, and deep infusion via multiple layers of knowledge introduction through attention mechanism at various levels. The attention layers enable the model to compute the significance score of words/phrases from user posts compared to the CPGs (here DepressionFeature and UMLS) and sentimental aspects (like COMET). It has three self-attention layers as shown in Figure \ref{fig:KiNN} which focuses on depression and sentiments. The first two layers independently find attention scores of phrase-tagged user posts related to depression concepts from DepressionFeature ontology (DFO) and emotional aspects from COMET. The third attention layer takes the concatenated output from the two and decides upon the importance of words/phrases compared to the depression and emotional aspects cumulatively.



{\fontfamily{cmss}\selectfont KiNN} is supported by two external modules: \textit{Knowledge Infusion from Domain-specific Knowledge Graphs}\footnote{The current architecture of {\fontfamily{cmss}\selectfont KiNN} uses two domain-specific knowledge graphs. Other forms of knowledge can also be explored.} and \textit{Explanation Visualizer}. Figure \ref{fig:KiNN} shows the architecture of {\fontfamily{cmss}\selectfont KiNN} with DepressionFeature Ontology and COMET transformer. User posts collected from social platforms, Reddit in this case, have been passed to the \textit{Phrase Tagging} module. The module employs the DepressionFeature Ontology to tag the domain-specific phrases from user posts.

\noindent \textbf{MentalBERT} \cite{ji2021mentalbert} is a BERT-based large language model pre-trained on mental health-specific data from social media platforms. BERT tokenizer, which considers words or sub-words as tokens, not phrases. Moreover, depending on the context, a single word could have more than one representation in the dimensional space. Hence, to calculate a fixed vector, phrases are considered to be present standalone, and representation from the last four layers of the encoder was added and averaged. Phrases are considered over words as tokens because phrase tagging preserves the meaning of sentences following the transformer tokenization process \cite{gu2021ucphrase}.   

\noindent The \textbf{DepressionFeature Ontology (DFO)} is a structured person and disease-specific data model to detect and monitor mental disorders in social media users'. It contains distinct features from multi-modal data regarding various mental disorders \cite{dalal2022featureonto, jain2023ontology}—for example, depression-related text phrases or image characteristics from social media users' postings. DepressionFeature ontology is populated with significant topics and static and dynamic-length phrases (uni, bi, tri, or quad grams) across all depressive posts in the PRIMATE, CLEF e-Risk, and CAMS datasets. Topics have been extracted through unsupervised LDA, while for phrases, besides the NLTK library's "ngram" function, they deployed three neural keyphrase extraction methods. These methods are KeyBERT \cite{grootendorst2020keybert}, KeyBART \cite{kulkarni2022learning}, and KeyBERT incorporating part-of-speech tagging and TF-IDF \cite{danesh2015sgrank}. DepressionFeature ontology also reuses instances from the Patient Health Questionnaire-9 Depression Ontology (PHQ-9 DO) \cite{dalal2023cross}. PHQ-9 DO has been developed to produce concept-level and user-understandable explanations where each category (or class in ontological terminology) represents a question from the PHQ-9. 

This work complemented the DepressionFeature ontology with another biomedical knowledge source, the \textbf{unified medical language system} (\textbf{UMLS}), for expert-curated synonyms and depression-related definitions of the phrases \cite{bodenreider2004unified}. Further, we have considered the \textbf{MentalBERT} \cite{ji2021mentalbert} to find similar (synonym or related) phrases used by social media users. Phrases in UMLS and MentalBERT are checked for cosine similarity with phrases from the ontology, and a threshold of cosine score of $=>0.80$ is considered to find the related or synonym phrases. All these resources (DepressionFeature ontology, UMLS, and MentalBERT) help identify and tag depressive phrases in user posts.

\noindent \textbf{COMET} \cite{bosselut2019comet}, a generative transformer model trained on ATOMIC \cite{sap2019atomic} is employed to extract everyday inferential if-then emotional aspects from user posts. This knowledge is present in the form of textual descriptions mainly focused on the speaker and listeners of the input. There are total nine different if-then aspects: i) \textit{the intent of W}, ii) \textit{the need of W}, iii) \textit{attribute of W}, iv) \textit{effect on W}, v) \textit{wanted by W}, vi) \textit{the reaction of W}, vii) \textit{effect on others}, viii) \textit{wanted by others}, ix) \textit{the reaction of others}. Here \textit{W} is the writer (or speaker of the post/comment) while others are listeners. As an example, with the input "Person \textit{W} gives him a compliment," \textit{the intent of W} would be \textit{W wanted to be nice}. Following the conviction of \cite{sap2019atomic, ghosal2020cosmic, zhu2021topic} who utilized these aspects in emotion and mental health-related tasks, we selected five aspects: \textit{intent of W} ($E_{IW}$), \textit{effect on W} ($E_{EW}$), \textit{reaction of W} ($E_{RW}$), \textit{effect on others} ($E_{EL}$), and \textit{reaction of others} ($E_{RL}$). Considering the close relation between emotions and mental states, this selection is expected to be appropriate.

\noindent \textbf{Explanation Generation} visualizes the attention patterns in {\fontfamily{cmss}\selectfont KiNN} against each attention block. The attention weights reflect the model's internal representation of the relationships. 
{\fontfamily{cmss}\selectfont KiNN} also represents dependencies between different words/phrases in the input and depression or sentimental aspects in the attention blocks. The visualizer identifies key parts of the input text regarding how the model processed the information and performed  \textit{depression classification.} The end-to-end training in {\fontfamily{cmss}\selectfont KiNN} generates a contextualized representation that any classifier can use. The current version of {\fontfamily{cmss}\selectfont KiNN} employs \textit{feed-forward neural network} (\textit{FFNN}) for classification.

\subsection{Algorithm}
Depending on the attention layers {\fontfamily{cmss}\selectfont KiNN} has two variants named {\fontfamily{cmss}\selectfont KiNN 1} and {\fontfamily{cmss}\selectfont KiNN 2}. {\fontfamily{cmss}\selectfont KiNN 1} has cross-attention for $\hat{u_{i}}\hat{E}$ while {\fontfamily{cmss}\selectfont KiNN 2} has self-attention network. Algorithm \ref{alg:KiNN_algorithm} presents the working process of {\fontfamily{cmss}\selectfont KiNN}. 

User posts in the form of paragraphs or sentences represented as $u_i$ are passed to {\fontfamily{cmss}\selectfont KiNN}, which outputs a contextualized representation $z_{u_i}$. Depending on the downstream task of detecting depression or causes of depression (see the  Datasets section), the representation is given to a \textit{feed-forward neural network} (\textit{FFNN}) classifier or the output layer. The process involves transforming $u_i$ into $u^{pt}_i$, where "$pt$" stands for phrase tagging. The phrases were identified and tagged using multiple domain-specific knowledge sources, as illustrated in Figure \ref{fig:KiNN}. Phrase tagging helps infuse depression knowledge in user posts. Domain-specific LM, MentalBERT \cite{ji2021mentalbert} embeds the user posts ($u^{pt}_i$) into a vector ($\hat{u}^{pt}_i$) to pass it through a self-attention layer. This attention layer focuses on significant parts of the post while considering the infused knowledge via tagged phrases.

In parallel, the emotional aspects along with commonsense are extracted for a user document ($u_i$) through pre-trained COMET on the ATOMIC knowledge graph. Each user document and corresponding text descriptions of five selected aspects ($E_{IW}$, $E_{EW}$, $E_{RW}$, $E_{EL}$, and $E_{RL}$) returned by COMET are concatenated ($u_{i}E$) and embedded using MentalBERT encoding ($\hat{u_{i}}\hat{E}$). This time, the user documents are not phrase tagged as the COMET is trained using BERT tokenizer, which considers words or sub-words as tokens, not phrases. The domain knowledge is infused with phrases; in mental health, phrases have more significance than words and sub-words. This significance is also visible in our previous studies \cite{dalal2023cross}.

\begin{algorithm}[t]
\caption{\textbf{\fontfamily{cmss}\selectfont KiNN} - Training Loop}
\label{alg:KiNN_algorithm}
\begin{algorithmic}[1]
\State I, E, R, W, and L stands for Intent, Effect, Reaction, Writer, and Listener
\While {loop over batch data samples $(u_i,y_i)$} \Comment $(u_i,y_i)$ is input pair indexed by $i$
\State Set input $u_i$ \Comment tokenized user document   
\State obtain $u_i^{pt}$ \Comment input after mapping to the Phrase Lexicon (i.e. DepressionFeature) (Run algorithm \ref{alg:UMLS_algorithm})
\State obtain $\hat{u}_i^{pt} = EL({u}_i^{pt})$ \Comment EL represents embedding layer (i.e. MentalBERT)

\For{$k$, where $k \gets 1~to~K$} \Comment {$K$ is no. of commonsense aspects in COMET}
\State $E_{I/E/R, W/L} = CL({u}_i)$          \Comment CL represents commonsense layer from COMET
\EndFor
\State obtain $u_iE = Con({u}_iE_{IW}E_{EW}E_{RW}E_{EL}E_{RL})$ \Comment Con represents concatenation layer
\State obtain $\hat{u}_i\hat{E} = EL({u}_iE)$ \Comment EL represents embedding layer (i.e. MentalBERT)
\State obtain $S_{\hat{u_{i}}^{pt}} = SA(\hat{u}_i^{pt})$ \Comment Self-Attention layer 
\State obtain $S_{\hat{u}_i\hat{E}} = SA(\hat{u}_i\hat{E})$ \Comment Self-Attention layer 
\State obtain $h_i = LN(Con(S_{\hat{u_{i}}^{pt}}S_{\hat{u}_i\hat{E}}))$ \Comment Concatenation and Layer normalization
\State obtain $S_{h_i} = SA(h_i)$ \Comment Self-Attention layer 
\State obtain $z_{u_i} = DL(h_i)$ \Comment Single Dense layer 
\State $\hat{y_i}(u_i) = \sigma(\Theta^T(\oplus ~z_{u_i}))$ \Comment Row concat and Predict
\State If $L(\hat{y_i}(u_i),y_i(u_i)) \leq \varepsilon$, break \Comment {Convergence check}
\EndWhile
\end{algorithmic}
\end{algorithm}

The embedded representation of $u_i$ and emotional aspects $\hat{u_{i}}\hat{E}$ is passed through the self-attention network. The network provides a significance score to each aspect according to their relevance to the user post. Further, the output from self-attention layers is merged and normalized before being transferred to a single attention layer to decide the relevance of domain-specific and emotional (commonsense) aspects. This way {\fontfamily{cmss}\selectfont KiNN} finds if a user document is mental health-related or general emotion (commonsense) related. Depending on the task,  hidden representation, $z_{u_i}$, of user posts generated by {\fontfamily{cmss}\selectfont KiNN} is fed into a classification head to obtain the final classification label or vector, $\hat{y}_i(u_i)$. We can interpret which aspects contributed the most predictive value toward the final classification by examining the attention matrices. Algorithm \ref{alg:KiNN_algorithm} provides a formal pseudocode for the complete process within {\fontfamily{cmss}\selectfont KiNN}. Notably, line 4 and 6 the associated domain knowledge source, which at present is DepressionFeature, can be changed.

Algorithm \ref{alg:UMLS_algorithm} discusses the process of mapping the user posts into DepressionFeature Ontology (DFO). Phrases from posts unavailable in DFO are searched in UMLS and the top three concepts are extracted and tagged in the corresponding user post. If UMLS also does not contain any phrase then  MentalBERT is employed to look for similar phrases. These will further be looked into UMLS for top concepts and tagged in user posts for processing by Algorithm \ref{alg:KiNN_algorithm}.

\begin{algorithm}[t]
\caption{\textbf{Lexicon Mapping}}
\label{alg:UMLS_algorithm}
\begin{algorithmic}[1]
\State Initialize DepressioFeature ontology (DFO) and UMLS
\While {loop over data sample $(u_i)$} \Comment $(u_i)$ is input user post indexed by $k$
\State Set input $u_i$ \Comment tokenized user document 
\State Tag tokens for Part of Speech to keep NOUNS only
\For{$W$, where $W \gets 1~to~l$} \Comment {$l$ is no. of noun tokens in user post $u_i$}
    \If{DFO has W}
        \State Tag $u_i$ with the matching phrase
    \ElsIf{DFO does not contain W}
        \State Find Synonyms and Similar tokens having similar contextual information (Employed MentalBert)  
        \If{DFO contains Synonym or Similar token}  
            \State Tag $u_i$ with the matching phrase
        \ElsIf{DFO does not have a Synonym and UMLS has W or its Synonym}
            \State Tag $u_i$ with the matching phrase
            \State Import top three Concept Unique Identifiers (CUIs) and their definition from SNOMED CT
        \EndIf
    \EndIf
\EndFor
\EndWhile
\end{algorithmic}
\end{algorithm}

\section{Experimental Setup}

\subsection{Datasets}
\noindent \textbf{CLEF e-Risk (Type: \textit{Binary}; Context: \textit{Depression}):} CLEF e-Risk sourced from r/depression subreddit consists of user posts and comments. We have considered the CLEF e-Risk dataset released in 2021. This version of CLEF is annotated with the Beck Depression Inventory (BDI), a CPG used by MHPs for detecting depression \cite{losada2016test, beck1987beck}. However, there is no existing lexicon for BDI, so we leverage an existing PHQ lexicon created by Yazdavar as the source of clinical groundedness \cite{yazdavar2017semi}. The dataset comprises at most 2000 Reddit posts per user for 828 users, out of which 79 users have self-reported clinical depression and 749 are control users. The control set consists of random Redditors interested in discussing depression.

\noindent \textbf{PRIMATE (Type: \textit{Multi-label}; Context: \textit{Depression}):}
The PRIMATE dataset was constructed to train conversational agents to judge which PHQ-9 questions are answerable from the user's online post. 
The dataset comprises 2,003 posts sourced from Reddit's r/depression\_help subreddit. Each post has been labeled with nine binary annotations that correspond to whether the post addresses one of the nine PHQ-9 items. The dataset has been established as a gold standard for assessing depression severity, demonstrating an inter-annotator agreement (using Fleiss Kappa \cite{sim2005kappa}) of 0.85 among six MHPs affiliated with the National Institute of Mental Health and Neurosciences (NIMHANS) in Bangalore, India. The anonymized dataset is made publicly available by \cite{gupta2022learning}.

\noindent \textbf{CAMS (Type: \textit{Multi-Class}; Context: \textit{Mixed Depression and Suicide}):} The CAMS dataset comprises 5051 instances and is designed to identify the primary causes of MH problems by categorizing social media posts into six causal classes \cite{garg2022cams}. These classes, which are based on the underlying reasons for mental illness and derived from relevant literature, include 'No reason' (C0), 'Bias or abuse' (C1), 'Jobs and careers' (C2), 'Medication' (C3), 'Relationship' (C4), and 'Alienation' (C5). The dataset is presented in a <text, cause, inference> format, where ‘text’ is the user post with 'cause' referring to the labeled reason behind the mental disorder mentioned in the post, and ‘inference’ indicates the key phrases (or relevant concepts) in the post that expert annotators considered when assigning labels. The annotations were reviewed by a clinical psychologist and a rehabilitation counselor and validated using Fleiss' Kappa interobserver agreement, achieving a substantial agreement of  0.61.

\subsection{Implementation and Training Details:}
While training {\fontfamily{cmss}\selectfont KiNN} on \textit{CLEF e-Risk Dataset} user post length is taken as 2000 i.e. a maximum of 2000 tokens (grams) are embedded for a user. Other parameters are two classes (Depressed vs. Non-Depressed), sixteen training and validation batches, fifteen training epochs, and a learning rate of one e-03. For \textit{PRIMATE dataset} the parameters were 9 classes, 150 user post length, 16 training and validation batches, 25 training epochs, and 1e-03 learning rate. Further, to train {\fontfamily{cmss}\selectfont KiNN} on \textit{CAMS Dataset} of 6 classes (C0-C5), maximum text length is taken as 50 with a training and validation batch size of 128. The training epochs were 25 and the learning rate is placed at 1e-03. The training is completed on an Intel Xeon server (8GB) which took about 8 hours. 


\subsection{Evaluation Metrics:}
Four evaluation metrics have been chosen to assess the effectiveness of {\fontfamily{cmss}\selectfont KiNN} across the three datasets. Quantitative evaluation is done using standard performance indicators such as Precision (P), Recall (R), F1, and Matthew Correlation Coefficient (MCC) scores. The creators of the original dataset employed these metrics to guarantee a fair benchmark and enable insightful comparisons. The micro average for P and R is found to be the same in most of the cases. So we have calculated the macro average for P, R, and F1 which provides equal weight to each class in multiple classes.

\section{Results \& Discussion}
Results from {\fontfamily{cmss}\selectfont KiNN} on multiple depression-related datasets are compared with popular generic and domain-specific transformers. This section presents qualitative results along with a user-level explanation generation process and results from the proposed model and one best baseline model.  

\subsection{Quantitative Results}
\textit{\textbf{{\fontfamily{cmss}\selectfont KiNN} on CLEF e-Risk Dataset:}} Table \ref{tab:KiNN_results} reveals consistent improvement from \textbf{KiNN} over baseline methods in the CLEF e-Risk dataset. In terms of precision, recall, and F1 score, {\fontfamily{cmss}\selectfont KiNN 2} outperforms all other models, followed by {\fontfamily{cmss}\selectfont KiNN 1}.
However, the MCC scores reveal a slightly different pattern. {\fontfamily{cmss}\selectfont KiNN 1} achieves the highest MCC score, indicating better overall classification performance in terms of true and false positives and negatives. In comparison with the best generic transformer model BERT, \textbf{{\fontfamily{cmss}\selectfont KiNN 2}} showed especially remarkable benefit, particularly in MCC (19\%) compared to precision (30.6\%), recall (41.6\%), and F1 (34.4\%) scores. However, while comparing \textbf{{\fontfamily{cmss}\selectfont KiNN 2}} with the second-best domain-specific transformer model, MentalBERT, significant enhancements were observed across various standard performance metrics. Specifically, in the case of MCC (25.2\%) compared precision (4.5\%), recall (20.5\%), and F1(13\%) scores. Similar results were obtained in the case of Naive Bayes (NB) and textCNN models. MCC is high for both the \textbf{KiNN} variants which signifies better overall performance of the classifier.




\textit{\textbf{{\fontfamily{cmss}\selectfont KiNN} on PRIMATE:}} 
For precision, recall, and F1 score, {\fontfamily{cmss}\selectfont KiNN 2} again emerges as the top-performing model, with \textbf{{\fontfamily{cmss}\selectfont KiNN 1}} following closely. Interestingly, in terms of MCC, {\fontfamily{cmss}\selectfont KiNN 1} achieves the highest score, indicating better overall classification performance. From generic transformer models, BERT gives higher scores than other models. {\fontfamily{cmss}\selectfont KiNN 2} has provided an approximate gain of 10\%, 14\%, 10\%, and 1\% over BERT in terms of precision, recall, F1, and MCC scores respectively. From domain-specific models MentalBERT second-best model after {\fontfamily{cmss}\selectfont KiNN}.




\textit{\textbf{{\fontfamily{cmss}\selectfont KiNN} on CAMS:}} provided ``inference,'' reported by annotators as an explanation behind their labeling. Such ground-truth explanations enable a more thorough assessment of the fidelity of the \textbf{\fontfamily{cmss}\selectfont KiNN}'s attention allocation. Table \ref{tab:KiNN_results} shows the performance gains of \textbf{{\fontfamily{cmss}\selectfont KiNN}} over CAMS baselines. 

\textbf{{\fontfamily{cmss}\selectfont KiNN:}} Though MentalBERT performed best in terms of F1 and MCC scores while BioBERT is best in precision and recall, {\fontfamily{cmss}\selectfont KiNN} variants demonstrated competitive performance across precision, recall, F1 score, and MCC. However, {\fontfamily{cmss}\selectfont KiNN 1} achieves the higher precision and F1 score on this dataset, while {\fontfamily{cmss}\selectfont KiNN 2} achieves the higher MCC.

\begin{table}[t]
\begin{center}
\scriptsize
\resizebox{8cm}{!}{
\begin{tabular}{p{2cm}|ccccc}
    \toprule[1.5pt]
    \multirow{2}{*}{\Large{\textbf{Models}}} &\multicolumn{4}{c}{\textbf{CLEF e-Risk} (in \%)}\\ \cmidrule{2-5}
    & \textbf{P} & \textbf{R} & \textbf{F1 }& \textbf{MCC}\\ \cmidrule{2-5}
    NB & 58.5 & 53.0 & 55.0 & 20.3 \\
    textCNN & 57.3 & 51.2 & 54.2 & 22.6\\ \cline{2-5}
    BERT & \underline{36.9} & \underline{34.4} & \underline{35.6}  & \underline{23.6} \\ 
    RoBERTa & 31.4 & 25.1 & 27.8 & 17.3\\
    MentalBERT & \underline{63.5} & \underline{56.5} & \underline{58.0} & \underline{18.8}\\
    BioBERT & 63.0 & 55.5 & 57.0 & 17.4 \\ \cline{2-5}
    \textbf{{\fontfamily{cmss}\selectfont KiNN 1}}$\dagger$ & \textbf{64.0} & \textbf{69.0} & \textbf{66.0} & \textbf{32.7} \\ 
    \textbf{{\fontfamily{cmss}\selectfont KiNN 2}}$\dagger$ & \textbf{67.5} & \textbf{76.0} & \textbf{70.0} & \textbf{42.6} \\
    
    \midrule[1pt]
    
    & \multicolumn{4}{c}{\textbf{PRIMATE} (in \%)} \\ \cmidrule{2-5}
    BERT & \underline{48.7} & \underline{42.6} & \underline{45.4} & \underline{13.8}\\
    RoBERTa & 44.2 & 41.5 & 42.8 & 11.4 \\
    MentalBERT & 54.0 & 35.0 & 37.0 & 12.4 \\
    BioBERT $\dagger$ & \underline{44.0} & \underline{36.0} & \underline{38.0} & \underline{6.5} \\ \cline{2-5}
   \textbf{{\fontfamily{cmss}\selectfont KiNN 1}}$\dagger$ & \textbf{53.0} & \textbf{52.0} & \textbf{52.0} & \textbf{19.8}\\
   \textbf{{\fontfamily{cmss}\selectfont KiNN 2}}$\dagger$ & \textbf{58.0} & \textbf{56.0} & \textbf{56.0} & \textbf{14.7}\\

    \midrule[1pt]
    
    & \multicolumn{4}{c}{\textbf{CAMS} (in \%)} \\ \cmidrule{2-5}
    MentalBERT & \underline{47.0} & \underline{43.0} & \underline{43.0} & \underline{33.8} \\
    BioBERT & 49.0 & 36.0 & 34.0 & 29.2 \\ \cline{2-5}
   \textbf{{\fontfamily{cmss}\selectfont KiNN 1}}$\dagger$ & \textbf{44.0} & \textbf{45} & \textbf{44.0} & \textbf{32.9} \\ 
   \textbf{{\fontfamily{cmss}\selectfont KiNN 2}}$\dagger$ & \textbf{41.0} & \textbf{41.0} & \textbf{41.0} & \textbf{29.1} \\ \bottomrule[1.5pt]
\end{tabular}}
\end{center}
\caption{\footnotesize\textbf{Classification results on CLEF e-Risk, PRIMATE, and CAMS.} The best performance is indicated in bold, while the second best is underlined. $\dagger$ (p $<$ 0.05) indicates statistically significant results when comparing the best to the second-best metric.}
\label{tab:KiNN_results}
\end{table}

\subsection{User-level Explanations}
For producing user-level explanations of depression classification results GPT 3.5 is employed. The "text-DaVinci-003" model is prompted in Python through the LangChain library \cite{Chase2023langchain}. The prompt is provided with user posts and significant words/phrases from MentalBERT or concepts from {\fontfamily{cmss}\selectfont KiNN}. Depression-related concepts are assigned higher attention scores by {\fontfamily{cmss}\selectfont KiNN} as compared to the MentalBert. GPT considers these concepts or words during explanation generation (see italicized words in Table \ref{tab:KiNN_GPT}). However, phrases from the MentalBERT model did not produce a comparable pattern in GPT 3.5's behavior. The reason could be observed from Figure \ref{fig:attentionvisualization} which displays highlighted text from MentalBERT and the proposed KiNN model. However, most of the time MentalBERT highlights all text with equal likelihood, and GPT 3.5 focuses on certain terms during the explanation-generating process.

\begin{table}[!t]
    \centering
    \scriptsize
    \begin{tabular}{p{8.5cm}}
    \begin{tcolorbox}[colback=white,
                  colframe=black,
                  width=8.5cm,
                  arc=1mm auto outer arc, title={Input Text},colbacktitle=white, coltitle=black
                 ]
         So, I decided to go to psychiatrist, but now I feel like I'm not that bad? How bad is really bad? Maybe I just act dramatic I cry a lot without any specific reason, I want to kill myself quite often, but... I dont know if I really need therapy? Maybe its just who I am, just a lazy whiner? And when I will come to therapist I dont know what to say, what if that day I will feel fine and they will not believe me? I dont look like I struggle all the time, people around me don't even know that I have problems, I've always been a comedian in a company. I'm just a little bit afraid. I dont know if i really need therapy, and if I don't, so what do I need to feel normal?
        \end{tcolorbox}\\ 
         \begin{tcolorbox}[colback=white,
                  colframe=black,
                  width=8.5cm,
                  arc=1mm, auto outer arc, title={MentalBERT on Words/Phrases},colbacktitle=white, coltitle=black
                 ]
        The text describes the narrator's inner struggle with determining if they truly need therapy or if they are just being dramatic. It also highlights their fear and uncertainty about seeking help and their desire to feel "normal". These concepts are connected to their decision to see a psychiatrist and their concerns about being perceived a certain way by others.
        \end{tcolorbox}\\ 
         \begin{tcolorbox}[colback=white,
                  colframe=black,
                  width=8.5cm,
                  arc=1mm, auto outer arc, title={{\fontfamily{cmss}\selectfont KiNN} on Concepts},colbacktitle=white, coltitle=black
                 ]
       The text discusses the narrator's decision to see a psychiatrist and their doubts about whether they truly \textit{need therapy}. The concepts identified relate to the narrator's mental state, including feelings of \textit{depression} (\textit{crying} and \textit{suicidal thoughts}), \textit{self-doubt} (worrying about being seen as a "\textit{lazy whiner}"), and the desire to \textit{feel "normal."} These concepts reflect a potential \textit{struggle with mental health} issues and the need for \textit{professional help}.
        \end{tcolorbox}
    \end{tabular}
 \caption{\footnotesize \textbf{Illustrative instances showcasing the quality of explanations from \textbf{KiNN} compared to ClinicalT5, employing GPT3.5.} {The explanations generated by \textbf{KiNN} exhibit an application-relevant explainability as they accurately capture the context within the user's expression by using concepts in DepressionFeature Ontology. Conversely, other GPT3.5(MentalBERT) explanations appear to deviate from the intended context.}}
    \label{tab:KiNN_GPT}
\end{table}

The method establishes a connection between attention words and concepts in DepressionFeature Ontology by computing their cosine similarity. The visualizer then extracts top-ranked concept(s) with a similarity greater than 0.80, considering them potential concepts for generating explanations using GPT 3.5. GPT 3.5 utilized the mapped concepts, as illustrated in Figure \ref{fig:attentionvisualization}, to generate an explanation, as shown in Table \ref{tab:KiNN_GPT}.

\section{Conclusion}
The BlackBox nature of AI models limits their practical implementation in the healthcare domain specifically in mental health assessment due to its subjective behavior. Banking upon a system that does not provide reasoning for its decisions could be too costly in the mental health domain costing someone's life. Further explanations produced by ad-hoc models are rarely understandable to end users. Present models miss domain knowledge and other aspects of a person's personality like commonsense or emotional. To handle the challenges, we presented \textbf{KiNN}, a deep knowledge-infused learning network that considers domain knowledge (e.g., DepressionFeature, UMLS) and multiple emotional aspects of users' text (COMET) infused through at multiple levels through attention networks to provide user-level explainability in critical domains like MH. 

The performance of our inherently explainable model is comparable to the latest Language Model Models (LLMs) trained on both generic and domain-specific datasets, including BERT, BioBERT, and MentalBERT. The model's performance on all three datasets can be found in Table 2 and Table 3. Moreover, the LLMs are trained on very large datasets and costly resources along with large time and space requirements compared to our model. 

The inclusion of clinical domain knowledge, such as Clinical Practice Guidelines (CPG), is crucial for the identification of various symptoms and experiences within unstructured text. LM algorithms can enhance their learning capabilities by incorporating a wider range of mental health (MH) targets. This expansion enables them to provide improved support for diagnosis, treatment, and intervention strategies.  

The KiNN model demonstrated superior performance in identifying relevant concepts in unstructured text compared to the baseline models. The infusion of knowledge in natural language processing for mental health involves the incorporation of logical reasoning, such as clinical practice guidelines and emotional aspects. This enables the AI model to engage in continuous learning, ensuring its robustness and resistance to errors. In addition, the establishment of a shared semantic understanding between human experts and language models facilitates collaborative decision-making.



\section*{Acknowledgment}
This work is supported by the University Grants Commission (UGC) of India. Any opinions, conclusions, or recommendations expressed in this material are those of the authors and do not necessarily reflect the views of the UGC.

%
%

\bibliography{manuscript}
\bibliographystyle{unsrt}

\end{document}